\documentclass{INTERSPEECH2023}
\usepackage{CJKutf8}
\usepackage{lipsum}
\usepackage[dvipsnames]{xcolor}

\renewenvironment{thebibliography}[1]{
  \begin{oldthebibliography}{#1}
    \setlength{\itemsep}{0em}
    \setlength{\parskip}{0em}
}
{
  \end{oldthebibliography}
}

\interspeechcameraready

\title{HK-LegiCoST: Leveraging Non-Verbatim Transcripts for Speech Translation}
\name{Cihan Xiao$^1$, Henry Li Xinyuan$^1$, Jinyi Yang$^1$, Dongji Gao$^1$, Matthew Wiesner$^{1,2}$, Kevin Duh$^{1,2}$, Sanjeev Khudanpur$^{1,2}$}
\address{
  $^1$Center for Language and Speech Processing, $^2$Human Language Technology Center of Excellence \\
  Johns Hopkins University, Baltimore, MD, USA
  }
\email{\{cxiao7, xli257, jyang126, dgao5, wiesner, khudanpur\}@jhu.edu, kevinduh@cs.jhu.edu}

\begin{document}

\maketitle
 
\begin{abstract}
%Speech translation has drawn significant interest in recent years due to its potential to facilitate cross-cultural communication and break down language barriers. While both speech recognition and (text) translation are poorly-defined tasks for languages without standardized writing conventions, speech translation provides a textual interface with primarily oral languages such as regional vernacular and dialects. 
We introduce HK-LegiCoST, a new three-way parallel corpus of Cantonese-English translations, containing 600+ hours of Cantonese audio, its standard traditional Chinese transcript, and English translation, segmented and aligned at the sentence level. We describe the notable challenges in corpus preparation: segmentation, alignment of long audio recordings, and sentence-level alignment with {\em non-verbatim transcripts}. Such transcripts make the corpus suitable for speech translation research when there are significant differences between the spoken and written forms of the source language.  Due to its large size, we are able to demonstrate competitive speech translation baselines on HK-LegiCoST and extend them to promising cross-corpus results on the FLEURS Cantonese subset. These results deliver insights into speech recognition and translation research in languages for which non-verbatim or ``noisy'' transcription is common due to various factors, including vernacular and dialectal speech.\footnote{The dataset is available at \url{https://huggingface.co/datasets/Borrison/hk-legicost}.}
\end{abstract}

\section{Introduction}
Growing demand for applications such as automatic video captioning and foreign language learning has spawned interest in improving speech-to-text translation (ST), which translates the speech of one language into the text of another language. While most work and progress has focused on high-resource languages, research on the translation of primarily spoken languages, or languages whose written forms deviate from their spoken forms, is relatively scarce. Cantonese is one such language, whose written form is often altered to appear closer to the Mandarin written form. We refer to this written form as standard Chinese and say these transcripts are "non-verbatim" to reflect the discrepancy between what is spoken and written.
%In particular, the lack of paired speech and textual translations makes it very difficult to apply the standard ST techniques which require a huge amount of manually prepared data.
%Cantonese and Mandarin differ in phonology, orthography, and grammar. However, Cantonese is often transcribed in standard written Chinese, which 
This presents significant challenges for automatic speech recognition (ASR) and speech translation (ST) systems.

In this paper, we introduce HK-LegiCoST: a new corpus of Cantonese audio recordings, corresponding standard Chinese transcripts, and paired English textual translation. It contains 600+ hrs of conversational and read speech collected by the Hong Kong Legislative Council~\cite{hklegco}, mainly focusing on government policy-related inquiries and their corresponding responses, alongside discussions and debates on motions and resolutions. 
%This 600+ hour corpus is sufficient for training standard ST systems.
We first describe the challenges in converting this raw ``found'' resource into a large, curated and useful corpus for language technology research.
We then provide automatic speech recognition (ASR), machine translation (MT), and speech translation (ST) baselines on this corpus. On the Google FLEURS ~\cite{fleurs} test set, an ASR model trained only on our training set leads to results that are comparable to Google's pre-trained and fine-tuned model results. Fine-tuning our baseline model on the FLEURS training set significantly outperforms the previously reported baseline. We also train and benchmark some standard speech translation models on our corpus. 
%Furthermore, the ST results with a BLEU score of 16.4 on the corpus's test set demonstrate the reliability of our corpus for speech translation research purposes. 
% The unique linguistic features of Cantonese, the transcription protocol using standard Chinese, and the available paired translations, and the relatively large amount of 3-way parallel data make this corpus a valuable subject for studying vernacular and dialectal speech recognition and translation.
We believe this corpus will become a valuable resource for studying vernacular and dialectal speech recognition and translation due to (a) the unique linguistic features of Cantonese, (b) the non-verbatim aspect of writing Cantonese in Standard Chinese, and (c) the relatively large amount of three-way parallel data.

Section~\ref{sec:related} provides an overview of prior research on speech translation and vernacular/dialectal speech recognition, Section~\ref{sec:creation} outlines the corpus creation pipeline, Section~\ref{sec:exp} presents our baseline experiments and their results, and Section~\ref{sec:discussion} explores some of the distinctive attributes of the corpus.
% \textcolor{red}{depends on how you organize your chapters (see Liz's paper for reference), it should include 
% \begin{itemize}
%     \item \textbf{a chapter of challenges} (long audio seg&align, how to decide it's well aligned, text align ...)
%     \item \textbf{a chapter of experiments} including a table of the details of the corpus (hours, lines, words), splits (train/dev/tests for ASR and ST); ASR exp description and results, ST description and results. The description of models can be summarized in tables if it takes too much space, you can create a table with important model parameters, i.e., layers, #BPEs, #nodes, ...; after each exp there should be a few lines of comments / conclusions
%     \item \textbf{a chapter of conclusions of this paper} 
% \end{itemize}
% }

\begin{figure}[!htbp]
  \centering
  \includegraphics[width=\linewidth]{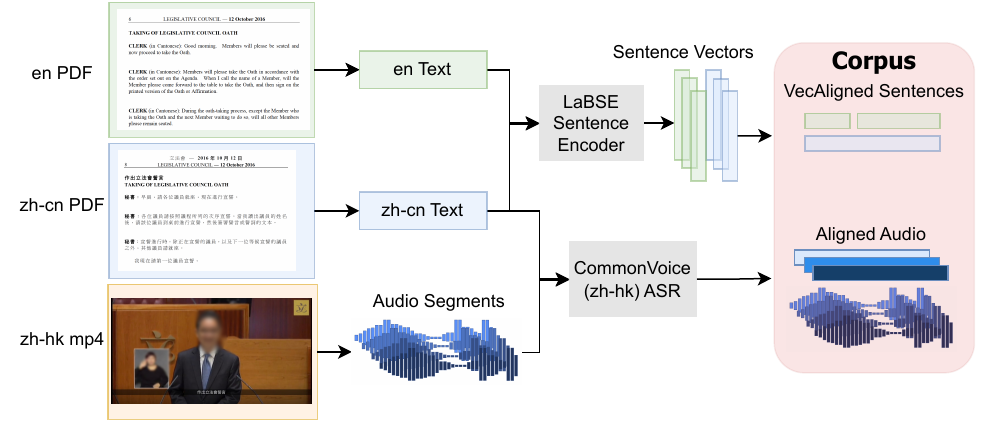}
  \caption{The corpus creation process. The standard Chinese text (zh-cn) is a non-verbatim transcript of the Cantonese speech (zh-hk).}
  \label{fig:pipeline}
\end{figure}

\section{Related work} \label{sec:related}

\subsection{Prior work on Speech Translation}
While historically the most competitive ST systems were cascade systems constructed by attaching an ASR system to an MT system, over the past couple of years greater focus has been placed on the growing potential of end-to-end ST systems. With that comes the need for ST corpora composed of audio recordings annotated with their transcripts translated into the target language(s). Work in this area kicked off with high resource language pairs such as Spanish-English~\cite{post-etal-2013-improved} or English-French~\cite{kocabiyikoglu-etal-2018-augmenting}, propelled by venues such as The International Conference on Spoken Language Translation (IWSLT). Multilinguality as well as coverage of lower resource languages soon followed: MuST-C~\cite{di-gangi-etal-2019-must} consisted of English speech aligned to transcripts in eight languages; more recent works such as CVSS~\cite{jia-etal-2022-cvss} and FLEURS~\cite{fleurs} saw an explosion both in the coverage of languages and in the quantity of data as measured in total speech length. FLEURS represents an important first step in that it was one of the earliest notable corpora for Cantonese speech translation; the size of its Cantonese portion is not sufficient for training a competitive speech translation system, a gap which our corpus aims to fill.
% and will use it as a reference benchmark for the baseline systems trained on our dataset.

Collecting, transcribing, and translating speech is a costly and difficult process, which has led works like Europarl-ST~\cite{europarl} and Multilingual TedX~\cite{mtedx} to leverage public-available, multilingual-captioned audio(visual) data, an approach that we emulate in our work. % Such an approach comes with its own set of challenges with cleaning and alignment, which we shall discuss in more detail in section~\ref{sec:creation}.

% \subsection{Cantonese ASR and MT corpora}
\subsection{ASR, MT and ST corpora for Cantonese, other vernaculars and/or unwritten languages}
% \subsection{Corpora for Cantonese, other vernaculars and/or unwritten languages}
% owing to Cantonese being the primary spoken language in Hong Kong
While corpora on many spoken forms of standard Chinese are few and far between, a plethora of corpora for spoken Cantonese have been developed. Yu et al.~\cite{yu-etal-2022-automatic} produced an excellent survey of existing Cantonese speech corpora, as well as a Cantonese ASR corpus comparable in size to the Cantonese section of the Common Voice corpus~\cite{ardila-etal-2020-common} or the Babel corpus~\cite{harper2011iarpa}. 

% All three are large enough for training a decently competitive ASR model or form part of the training data for a multilingual ASR system, but fall short of what is needed for performances comparable to models trained on high-resource language pairs.

Several other studies have drawn from the same data source -- the Legislative Council (LegCo) meeting records -- for translation-related studies. One of the most notable corpora for Cantonese MT is the Hong Kong Hansards Parallel Text \cite{ma2000a}, which draws from LegCo's records between 1995 and 2000. Kwong's 2021 study \cite{Kwong2021} extracted a small number of sentence pairs from LegCo meeting records to conduct a focused analysis of the linguistic characteristics of translation and interpretation of Cantonese. By contrast, we aim to support cutting-edge research in ASR and ST using a large amount of resources.

% The source material of our dataset, the LegCo meeting parallel transcripts, was used in an earlier study\cite{Kwong2021} which conducted a focused study on the linguistic characteristics of the translation and interpretation in those transcripts. Only a relatively small number of sample sentence pairs were extracted in that study, which contrasts with our objective of leveraging this data to provide as many resources as possible for supporting cutting-edge research in ASR and ST.

% \subsection{ASR and ST corpora for vernaculars and/or Unwritten Languages}

%While sporadic compared to speech corpora in high-resource languages, 
Vernacular speech corpora have grown in number in the recent past.
%and provide a unique opportunity for speech researchers.
The MGB-2 Arabic dataset \cite{mgb2} has served as the backbone for the eponymous MGB-2 challenge, and contained some amount of dialectal Arabic with non-verbatim transcripts. Subsequent challenges, such as the IWSLT Tunisian Arabic speech translation challenge \cite{antonios2022findings}, focused specifically on dialectal translation, but the transcripts were verbatim. SDS-200 \cite{pluss-etal-2022-sds} is a recent work that tackles the speech translation between Swiss German and Standard German. 
% Other notable works on vernacular speech translation include the MaDiTS corpus \cite{Khaw2014PreparationOM} for Kelantanese Malay and the Thai-Isan parallel corpus by Seresangtakul and Unlee \cite{isan}.

One important characteristic of Cantonese is that it is a partially unwritten language: while it has its own writing system that captures the phonological and phonetic characteristics of spoken Cantonese, there are many situations where it is instead transcribed in standard written Chinese which does not have this property\cite{cantonese_tutorial}. The source material of our dataset falls into the latter category. A recent study by Chen et al. \cite{unwritten} addressed a task in a similar setting (Hokkien ST), although both works and data in this area are still relatively scarce.

\section{Corpus Creation} \label{sec:creation}

\subsection{Data Collection and Preprocessing}
The raw data of the corpus is a collection of video recordings of the Hong Kong Legislative Council's regular meetings, their corresponding transcripts, and English translations in PDF format. The Hong Kong Legislative Council, also known as HKLegCo, is the unicameral legislature of the Hong Kong Special Administrative Region of China.
%Its primary role is to enact, amend, or repeal laws, approve budgets, and monitor the work of the government.
The corpus was mainly derived from the recordings of the council's regular meetings from 2016 to 2021, covering a wide range of topics such as political reform, education policy, housing issues, transportation infrastructure, healthcare reform, and economic development.

To process this raw data into a usable corpus, we developed an integrated pipeline shown in Figure~\ref{fig:pipeline}.\footnote{The code of the pipeline is available at:
% \url{https://anonymous.4open.science/r/espnet-3FA1/egs2/commonvoice\_align/asr1/align.sh}}
\url{https://github.com/BorrisonXiao/espnet-align/tree/master/egs2/commonvoice\_align/asr1/align.sh}} It accepts the original video recordings of each meeting, and the corresponding transcript and English translation in PDF format, and produces triplets of segmented sentence-sized audio clips, together with their transcript and translation. Each module in the integrated pipeline entails one or more technological tasks. In the following sections, we provide a comprehensive description of the technological steps involved in each of the tasks.

% from 11:00 am to 7:00 pm, with a lunch break in between.
Typically, a regular meeting of the Council lasts for about half a day, with a video released for each meeting session. We start off by converting each video recording to WAV format, resampling it to 16KHz, and segmenting it based on metadata, i.e. the topic-level timestamps associated with each video indicating the start time of a certain topic (or sometimes speaker), yielding audio clips whose length generally spans from 5 to 30 minutes. As the raw recordings contain visual information such as the speaker's lip movements and sign language that corresponds to the speech, we plan to release a version of the corpus with visual information in a subsequent iteration of this work. %Figure~\ref{fig:pipeline} presents a summary of the integrated pipeline we built for corpus creation. 

% using the python module fitz
% Owing to the nature of meeting minutes, supporting information such as supplement documents, clarifications, charts, and figures are often present in these transcripts without being part of the audio. To address this, we extracted only paragraphs preceded by a speaker ID and a colon in the bolded format.
% A byproduct of such an extraction criterion is the bilingual textual alignment of these text segments, as the speakers made speeches in the same order across both transcripts.
The goal of text preprocessing is to filter out irrelevant information in the PDF transcripts and segment the full document into shorter sections. The raw text was first extracted from meeting transcripts and translation files. Paragraphs with a Chinese speech marker (name plus colon in bolded) are extracted from the raw text. This process automatically divides the full document into speaker ID labeled segments, allowing for more efficient bitext and audio-text alignment. 

\subsection{Bitext Sentence Alignment}
% , as it helps to ensure that the resulting bitext captures sufficient context for accurate translation
Sentence-level bitext alignment is crucial for ST corpus creation. To this end, we split each speaker-ID-labeled text segment into sentences based on punctuations, and generate a contextualized sentence embedding for each sentence using LaBSE~\cite{labse}, a multilingual sentence BERT model. We then used VecAlign~\cite{vecalign} to perform alignment on the embeddings. The VecAlign algorithm works by using pre-trained sentence embeddings to compute a similarity score between sentences in the source and target languages, followed by applying a dynamic programming approach based on the Fast Dynamic Time Warping algorithm to approximately find the optimal alignment between the sentences in linear time. We utilized the alignment score generated by VecAlign to identify and eliminate inaccurate alignment results (score $> 0.627$). The threshold was obtained via a grid search with manual inspection of the filtered instances.
% We observed that many of the excluded alignment instances were associated with formatting-related errors.

\subsection{Audio-Text Sentence Alignment} \label{subsection:a2t}
% 1. Write about the ASR acoustic model setting and the Cantonese-ASR vs Cantonese-Chinese-ASR issue (motivates the selection of the cv data). Also mention the vocabulary setting.

\subsubsection{Alignment Model Training}
% One of the main challenges we faced during audio-text alignment was the presence of transcript mismatch introduced by the standard written Chinese transcripts. To mitigate this issue, w
We performed audio-text alignment using an ASR model trained on the Cantonese language subset of the CommonVoice corpus~\cite{commonvoice:2020} with the ESPNet toolkit~\cite{watanabe2018espnet}. Specifically, the model used an encoder-decoder architecture with a Conformer~\cite{conformer} encoder, which was trained under the CTC/attention hybrid multitask learning~\cite{ctc_attention_mtl} framework without precomputed feature. One of the challenges of alignment model training is the discrepancy between what is written and what is spoken, i.e. non-verbatim transcripts. Therefore, we chose the CommonVoice corpus as our training data for the alignment ASR model, since its transcripts are similar to standard written Chinese. 
%than other conversational Cantonese ASR corpora such as IARPA Babel~\cite{harper2011iarpa}.
To reduce the disparity in vocabulary, we adopted a character-based tokenization approach for Chinese characters, while code-mixed English words and numbers were treated as single characters during the tokenization process. We converted the characters into jyutping, a romanization system for Cantonese, with the python-pinyin-jyutping-sentence tool,\footnote{\url{https://github.com/Language-Tools/pinyin-jyutping}} for addressing the vocabulary differences between the training and target corpora. We also used k2,\footnote{\url{https://github.com/k2-fsa/k2}} a toolkit for integrating Finite State Automaton (FSA) and Finite State Transducer (FST) algorithms with autograd-based machine learning models, to construct WFST-based lattice and perform decoding.

\subsubsection{First-pass (Paragraph) Alignment}
% 2. Write about the first pass alignment process, VAD + biased LM decoding + document-level textual alignment + anchor-based approach.
We first conducted a first-pass alignment process that approximately aligns audio segments with corresponding sections in the transcript using an anchor-based method. To achieve this, we utilized Silero-VAD~\cite{SileroVAD}, a voice activity detection tool, to extract audio segments of suitable duration for decoding. Next, we created a 3-gram language model that was biased towards the target document. Using this biased language model as G graph, we applied HLG decoding to the VAD-segmented audio clips. After decoding, we concatenated the decoding results for audio clips from the same meeting and aligned them with the full transcript text using KALDI's align-text tool~\cite{kaldi}. We identified anchors, which represent reliably aligned regions of the original audio segment, based on the alignment produced by SailAlign~\cite{sailalign}, using a set of criteria including CER, number of consecutive errors, and absolute errors. We expanded the boundaries of these anchors to account for potential decoding errors and mapped the resulting reference text regions to the audio clip, generating the first-pass alignment output.

\subsubsection{Sentence-level Alignment}
% 3. Write about the sentence-level alignment problem and the OOM problem, describe the sliding-window-based flexible alignment algorithm and how it solves the issues.

% despite our efforts to extract and filter the text beforehand, 
% Upon manual inspection, it was observed that the missing text is at the sentence level, indicating that a sentence in the transcript is either spoken completely or not at all.
In our efforts to perform sentence-level alignments, we encountered two primary challenges. Firstly, we found that supplementary text that was not spoken during the meeting was present in the transcript, complicating the alignment process. Secondly, for longer audio segments lasting upwards of 10 minutes, decoding proved to be challenging due to memory constraints. To overcome these challenges, we implemented a sliding-window flexible alignment algorithm, which enabled us to address both issues by efficiently breaking down long segments into smaller ones of manageable lengths while simultaneously filtering out non-spoken text from the transcript.

The basic flexible alignment algorithm is a variation of the forced-alignment algorithm.\footnote{
\url{https://github.com/DongjiGao/flexible\_alignment.git}}
% \url{https://anonymous.4open.science/r/flexible\_alignment-3AD7}}
Specifically, the algorithm employs a linear FSA as the G graph for decoding, but includes a skip connection from the start of a sentence towards the end of the sentence with customizable weight. This approach allows the algorithm to skip a sentence that may contain irrelevant text and focus on the next sentence that is more likely to be spoken. Our approach resembles the factor transducer~\cite{factor_transducer} method, which enables the decoding of sub-strings for long audio alignment by adding a skip connection from the start state to each state in the linear FSA and making each state a terminal state. However, unlike the factor transducer, our approach operates at the sentence level and permits the skipping of redundant sentences in-between.

\begin{figure}[t]
  \centering
  \includegraphics[width=\linewidth]{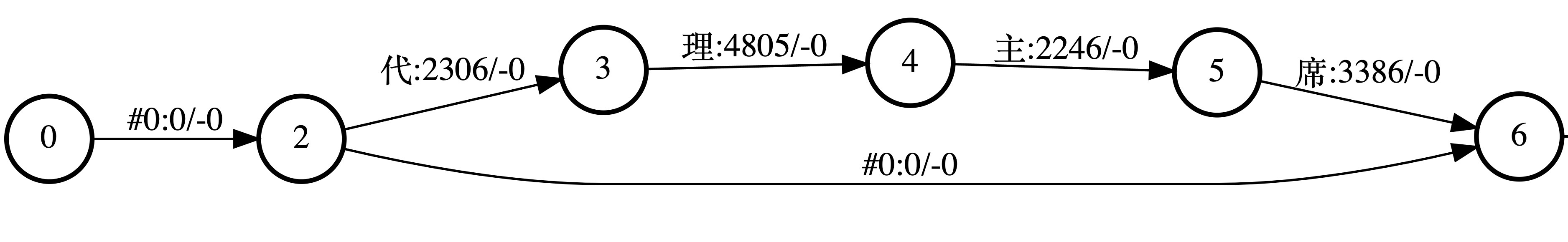}
  \caption{A sample G graph for flexible alignment.}
  \label{fig:flex_graph}
\end{figure}

% Factor transducer for alignment, substring decoding, comparison, related
% The output of the algorithm is a set of sentence-level time boundaries that align each sentence in the transcript with its corresponding audio segment. 
To further improve the efficiency and performance of the algorithm, a sliding-window strategy is used to break down audio segments into shorter, overlapping sub-segments, which can be processed in parallel on a GPU. The length of the decoding graph decreased with the length of the input audio signal, leading to a reduction of the search space and therefore a significant improvement in the alignment accuracy. The algorithm has been shown to effectively filter out unspoken text and improve the accuracy of sentence-level alignments, particularly for longer audio segments. 

% Overall, the sliding-window flexible alignment algorithm is a promising tool for enhancing the quality and efficiency of audio-text alignments in speech recognition and translation corpora.

\subsubsection{Post-Filtering}
% 4. Write about the ASR-based post-filtering and the WER + heuristic criteria, mention also about the human annotation.
In our pipeline, we utilized ASR decoding to evaluate the quality of the alignment by measuring the CER, number of consecutive errors, and error ratio. To ensure optimal alignment results, we established a threshold by creating CER-based bins and randomly sampling 300 utterances from the corpus based on the bins. We manually labeled the samples and fine-tuned the threshold to optimize the precision of the subset and to filter out instances where the speech contained significant deviations from the transcript due to repetitions and other disfluencies.
% We conducted an error analysis and found that the discrepancies between the speech that contains repetitions and other disfluencies and the transcript that was intentionally cleaned by the annotator for concision and formality contributed significantly to the errors. 
% As a result, we excluded such instances to enhance the quality of the corpus.
% To improve the quality of the corpus, we also tuned the criteria to filtered out instances where the speech contained significant deviations from the transcript, due to repetitions and other disfluencies.

% TODO: Mention the segmentation is a research topic which is why

\begin{table}[th]
  \caption{Split Overview. The perplexity of the dev and test sets is computed based on a 3-gram language model trained on the training split. The LM used is character-level for Cantonese and word-level for English.}
  \label{tab:split}
  \centering
  \addtolength{\tabcolsep}{-1pt}
  \small
  \begin{tabular}{ l r r l@{}r c@{}c }
    \toprule
    \multicolumn{1}{l}{\textbf{Split}} & \multicolumn{1}{r}{\textbf{Hours}} & \multicolumn{1}{r}{\textbf{\# Utts~~~}} & \multicolumn{2}{c}{\textbf{\# Tokens}} & \multicolumn{2}{c}{\textbf{Perplexity}}\\
    & & & zh-hk~~~ & en & zh-hk~~~ & en\\
    \midrule
    train       & $518$   & 142K~~~   & 6.8M~~~ & 4.8M & - & - \\
    dev-ASR     & $41$    & 11.1K~~~   & 557K~~~ & 380K & 36.7 & 394.8 \\
    % dev-asr-0   & $12.45$    & 3.3k~~~    & 167k~~~ & 114k      \\
    % dev-asr-1   & $12.57$    & 3.3k~~~    & 166k~~~ & 114k       \\
    % dev-asr-2   & $16.73$    & 4.4k~~~    & 224k~~~ & 152k       \\
    dev-MT      & $40$    & 10.7K~~~    & 519K~~~ & 371K & 39.5 & 371.5 \\
    % dev-mt-0    & $11.76$    & 3.2k~~~   & 152k~~~ & 109k\\
    % dev-mt-1    & $11.61$    & 3.2k~~~   & 156k~~~ & 111k\\
    % dev-mt-2    & $16.15$    & 4.2k~~~   & 211k~~~ & 150k\\
    test        & $10$     & 2.5K~~~   & 123K~~~ & 84K &40.3 & 471.5 \\
    \midrule
    \textbf{Total} & $609$ & 166.3K~~~ & 8M~~~ & 5.6M & - & - \\
    \bottomrule
  \end{tabular}
\end{table}

\subsection{Corpus Splits}
% ASR, speaker disjoint, MT document disjoint, motivation, ST both disjoint, matching distribution is hard
%We used NACHOS,\footnote{\url{https://anonymous.4open.science/r/nachos-DDF2}} a tool that automates the creation of heldout partitions using metadata of the units to be split, to generate the training, development, and test splits. 
We partitioned the data into four major subsets: training (train), ASR development (dev-ASR), MT development (dev-MT), and test (test) sets. We ensured that the test set was disjoint in both documents and speakers with respect to the training set, while preserving the gender distribution in the training set. 
The splits were created using \texttt{nachos},\footnote{
% \url{https://anonymous.4open.science/r/nachos-DDF2}}
\url{https://github.com/m-wiesner/nachos.git}}
a tool that automates the creation of held-out partitions using metadata of the units to be split.
Specifically, we defined each document in the corpus as the transcript of a full meeting recording, and used the document ID, speaker ID, and speaker gender as features for \texttt{nachos}. As a result, the dev-ASR set was created with speaker disjointness from the train set, while the dev-MT set was formed with document disjointness from the train set. The test set was both speaker and document disjoint from the train set. In addition, NACHOS approximately preserved the gender and speaker distribution across all splits based on gender features. Table~\ref{tab:split} displays the statistics of the splits. The perplexity measure suggests, as expected, that the Cantonese content in the dev-MT and test split is less similar with respect to the training set comparing to the dev-ASR split. Yet, the same pattern does not apply for the English text, presumably due to the translator switches across documents. In addition, it is worth mentioning that we partitioned the dev-ASR and dev-MT into three subsets each to reduce the training and validation cost.

\section{Baseline Experiments and Results} \label{sec:exp}
\interfootnotelinepenalty=10000

\subsection{Speech Recognition} \label{sec:asr}
% Trained from scratch baseline Conformer
We employed icefall to create baseline ASR models for the corpus.\footnote{\url{https://github.com/k2-fsa/icefall}} The 100M parameters Conformer model with CTC loss was trained using the same character-level tokenization and pronunciation lexicons as detailed in Section \ref{subsection:a2t}.\footnote{The code for model training and finetuning is available at 
% \url{https://anonymous.4open.science/r/icefall-880C/egs/hklegco/ASR/train.sh}}
\url{https://github.com/BorrisonXiao/icefall/blob/master/egs/hklegco/ASR/train.sh}}
 80-dimensional fbank features were used as input features. We trained a baseline model using only the HK-LegiCoST data and then finetuned it for 20 epochs on the FLEURS training set.

%We converted the decoded hypotheses from traditional Chinese to simplified Chinese to compare them with the reference transcript.
To further evaluate the robustness of the corpus, we tested the model on the yue\_hant\_hk subset of Google's FLEURS corpus which, similar to our dataset, contains Cantonese speech transcribed in standard simplified Chinese. We evaluated the model's zero-shot performance and performance after fine-tuning on the FLEURS training set. Despite being only 1/6th the size of the FLEURS baseline model, the HK-LegiCoST baseline model achieved competitive results in the zero-shot setting, while the fine-tuned model significantly outperformed the FLEURS baseline. Furthermore, the fine-tuned model suffered no performance degradation on the HK-LegiCoST test set, which indicates that the two corpora share a similar domain. These results demonstrate the robustness of our corpus from the ASR perspective.
% TODO: Push code
% TODO: Test Chinese pre-trained models, but in order to compare maybe need a larger baseline?
\begin{table}[th]
  \caption{CER$\downarrow$ of the baselines. The Conformer model is the baseline trained on the HK-LegiCoST training corpus, while the Conformer + FT model represents the finetuned version of the Conformer baseline on the FLEURS training set.}
  \label{tab:baselines}
  \centering
  \addtolength{\tabcolsep}{-1pt}
  \small
  \begin{tabular}{ l r r }
    \toprule
    \multicolumn{1}{c}{\textbf{Model}} & \multicolumn{1}{c}{\textbf{HK-LegiCoST}} & \multicolumn{1}{c}{\textbf{FLEURS}} \\
    \midrule
    FLEURS~\cite{fleurs} & - & $37.0$ \\
    Conformer & $23.2$ & $42.9$ \\
    Conformer + FT & $\mathbf{23.0}$ & $\mathbf{26.3}$ \\
    \bottomrule
  \end{tabular}
\end{table}

\subsection{Machine Translation} \label{sec:mt}

Our baseline MT system was built using FAIRSEQ~\cite{ott-etal-2019-fairseq}, employing a transformer-based architecture~\cite{NIPS2017_3f5ee243}. We adopted a unigram character-level tokenization method to tokenize the source (Cantonese) text and a BPE-based~\cite{sennrich-etal-2016-neural} approach to tokenize the target English text. Our baseline achieved a BLEU score of 24.9 on our test set. Table~\ref{tab:mtst} compares our MT system with one of the best existing benchmarks, M2M-100~\cite{m2m100}. %The column titles "unigram" and "bpe"~\cite{sennrich-etal-2016-neural} represent the method of tokenization we employed for the input (Cantonese) data; bpe was used to tokenize the target (English) data. 
Zero-shot M2M100 performed very poorly on named entities, whereas our system performed much better in that regard; this, combined with the relatively large number of similar utterances in parliamentary proceedings, resulted in our trained-from-scratch system pulling a sizeable gap ahead of M2M100. %despite having been trained on a smaller amount of data.
% In order to match the training input to what our model would get in the cascaded setup, punctuations were removed from the source sentences during training.
%, especially when it comes to the names of the people referenced in our corpus due to the difference in romanization standards between Mandarin Chinese and Cantonese

\begin{table}[th]
  \caption{BLEU$\uparrow$ of MT and Cascaded ST on HK-LegiCoST test set.}
  \label{tab:mtst}
  \centering
  \addtolength{\tabcolsep}{-1pt}
  \small
  \begin{tabular}{ l r r }
    \toprule
    \multicolumn{1}{c}{\textbf{Input}} & \multicolumn{1}{c}{\textbf{Ours}}~~ & \multicolumn{1}{c}{\textbf{M2M100}}  \\
    \midrule
    Oracle transcript & $\mathbf{24.9}$~~~ & $12.1$~~~ \\
    ASR hypotheses & $\mathbf{17.3}$~~~ & $7.9$~~~ \\
    \bottomrule
  \end{tabular}
\end{table}

\subsection{Cascaded Speech Translation}
We report the BLEU scores of the two cascaded systems, each of which takes the output of the ASR system as described in section~\ref{sec:asr} and translates with the two MT systems outlined in section~\ref{sec:mt}. The results are shown in Table~\ref{tab:mtst}. The unigram-input transformer vastly outperformed the bpe-input pre-trained m2m-100 model: the latter suffered greatly from the mismatch between its input vocabulary and the output vocabulary of our ASR system.

% The results of the ST experiments indicate that the cascaded system with MT model trained from scratch outperforms the system that uses the pre-trained m2m-100 model in its zero-shot performance on the test set. It is our belief that this performance gap is largely due to the fact that m2m-100 is dependent on word-level segmentable text, whereas the ASR output is at the character-level. This assertion is supported by the experiment results, as the ST performance using the unigram tokenized MT model significantly outperforms that of the BPE tokenized MT model.
% Experiments show that the models trained from scratch perform better than the pre-trained m2m-100 models' zero-shot performance on the test set. We conducted an error analysis, which revealed that the performance gap was largely caused by the reference documents' preference towards certain politically-related terms and unique named entities, as well as the speakers and translators' patterned use of particular lexcial items. 

\section{Discussion} \label{sec:discussion}
We observed two notable features in the proposed corpus, namely the phenomenon of word and phrase reordering introduced by non-verbatim transcript and the frequent occurrence of document-level contextual dependencies such as co-reference and named entities.

\subsection{Text Reordering}
\begin{CJK*}{UTF8}{bsmi}
Two major factors were identified as contributing to the phenomenon of word and phrase reordering. 
% The first factor was the discrepancy between Cantonese and standard Chinese. As a colloquial variant of the Chinese language family, Cantonese possesses its own vocabulary, grammar, and pronunciation distinct from other branches of Chinese~\cite{cantonese_tutorial}. 
The difference between the grammar of Cantonese and standard written Chinese can lead to word reordering. For instance, the phrase "你走先" (you go first) in Cantonese, when transcribed to standard Chinese, is converted to "你先走" (you first go). In addition, the formalization of the speech, as illustrated in table \ref{tab:formalization}, was an outcome of the transcriber's attempt to enhance the fluency of the text in standard Chinese.
\begin{table}[th]
  \caption{An example of data triplet in the corpus, where the differences between Cantonese and standard written Chinese are highlighted in blue, the results of lexical replacement for linguistic fluidity enhancement are highlighted in red, and the reordered text resulting from formalization is underlined.}
  \label{tab:formalization}
  \centering
  \begin{tabular}{p{0.35\linewidth} p{0.6\linewidth}}
    \toprule
    Cantonese Speech & \textcolor{blue}{佢係}我在上水石湖墟辦事處\underline{經常\textcolor{red}{來}}的街坊\\
    % \midrule
    Jyutping & \textcolor{blue}{keoi5 hai6} ngo5 zoi6 soeng5 seoi2 sek6 wu4 heoi1 baan6 si6 cyu3 \underline{ging1 soeng4 \textcolor{red}{loi4}} dik1 gaai1 fong1\\
    \midrule
    Reference Transcript & \textcolor{blue}{她是}\underline{經常\textcolor{red}{到訪}}我在上水石湖墟的辦事處的街坊 \\
    Jyutping & \textcolor{blue}{taa1 si6} \underline{ging1 soeng4 \textcolor{red}{dou3 fong2}} ngo5 zoi6 soeng5 seoi2 sek6 wu4 heoi1 dik1 baan6 si6 cyu3 dik1 gaai1 fong1\\
    \midrule
    English Translation & She is a local resident who has made frequent visits to my office in Shek Wu Hui in Sheung Shui\\
    \bottomrule
  \end{tabular}
\end{table}
\end{CJK*}
% 她是我在上水石湖墟辦事處經常来的街坊
% 她是經常到訪我在上水石湖墟的辦事處的街坊
% She is a local resident who has made frequent visits to my office in Shek Wu Hui in Sheung Shui

% \subsection{Code Mixing}
% Cantonese, owing to its historical development in a multilingual context, features a lexicon that includes numerous transliterations and is marked by code-mixing among speakers, particularly with regard to English terms. This poses a challenge to speech recognition systems, which must accurately recognize each language or variety being used. Our experiments demonstrate that in cascaded systems, failure of the ASR module to correctly recognize code-switched utterances frequently results in error propagation to translations. However, we contend that if speech signals are employed effectively, such as in a well-designed end-to-end ST system, code-mixing, particularly the use of target language words, may actually enhance translation quality.
% More statistics, per-utterance WER
% Tables with more numbers about different (extra) language models

\subsection{Long Context Dependency}
Due to the nature of the council meetings, it is common for the translations of the transcripts to exhibit contextual dependencies at the document level.
% This phenomenon is present in our corpus: during spontaneous exchanges, parliamentary members often assume knowledge of the meeting agenda as well as the meeting proceedings leading up to their statement, the knowledge that a sentence-segmented translation system wouldn't have access to. 
Errors made by our baseline translation system are often attributed to instances of such contextual dependence. We speculate that a translation system that could make use of these types of contextual information would outperform our baseline system.

\section{Conclusions}
% We are going through the formality of releasing the dataset, etc.
% Include link to the hk website
We created the HK-LegiCoST corpus for the research of vernacular speech recognition and translation. The original data was collected from publicly-available meeting recordings and transcripts from the Hong Kong Legislative Council. Our corpus, with 518 hours of Cantonese speech and 142k sentence pairs, is (to our knowledge) one of the largest for both Cantonese ASR and Cantonese-English speech translation. We report some of the linguistic characteristics in our corpus, namely text reordering and long context dependency. Results from our baseline experiments show that our corpus is validated and of great value for the research of ASR and ST. In addition, our approaches successfully address some of the top challenges for creating a corpus from scratch, especially for segmenting and aligning long recordings with non-verbatim transcripts.

\section{Acknowledgements}
% \textcolor{red}{Here we need to think whether / how to mention HKLegco}

% \ifinterspeechfinal
%      The INTERSPEECH 2023 organisers
% \else
%      % Neha Verma and Weiting Tan
% \fi

We are grateful to the Legislative Council of the Hong Kong Special Administrative Region for making their meeting proceedings and bilingual transcripts publicly available, which has allowed us to use them as a resource for academic research. We thank Neha Verma and Weiting Tan from Johns Hopkins University for helpful early discussions.
% would like to thank ISCA and the organising committees of past INTERSPEECH conferences for their help and for kindly providing the previous version of this template.

% We would like to thank Neha Verma and Weiting Tan for their helpful technical advice.

\bibliographystyle{IEEEtran}
\small
\bibliography{mybib}

\end{document}